\documentclass[11pt]{article}

\usepackage[preprint]{acl}

\usepackage{times}
\usepackage{latexsym}
\usepackage{bm}
\usepackage[T1]{fontenc}

\usepackage[utf8]{inputenc}
\usepackage{enumitem}

\usepackage{microtype}

\usepackage{inconsolata}

\usepackage{booktabs} 
\usepackage{graphicx}
\usepackage{tabularx}
\usepackage{array}
\usepackage{amsmath, amssymb}
\usepackage{algpseudocode}
\usepackage{algorithm}
\usepackage{multirow}
\usepackage{makecell}
\usepackage{pifont}
\usepackage{soul}
\usepackage{hyperref}
\usepackage[normalem]{ulem}

\usepackage{textgreek}
\usepackage[textsize=scriptsize]{todonotes}

\title{Memorization vs. Reasoning: Updating LLMs with New Knowledge}

\author{Aochong Oliver Li \\ Computer Science, Cornell University \\
  \texttt{aochongli@cs.cornell.edu } \\\And
  Tanya Goyal \\ Computer Science, Cornell University \\
  \texttt{tanyagoyal@cornell.edu} \\}

\newcommand{\method}{KUP} 
\newcommand{\evaltool}{KUPEval}
\newcommand{\mrl}{MCT}

\begin{document}
\maketitle

\begin{abstract}
Large language models (LLMs) encode vast amounts of pre-trained knowledge in their parameters, but updating them as real-world information evolves remains a challenge. Existing methodologies and benchmarks primarily target entity substitutions, failing to capture the full breadth of complex real-world dynamics. In this paper, we introduce \textbf{Knowledge Update Playground} (\method), an automatic pipeline for simulating realistic knowledge updates reflected in an evidence corpora. KUP's evaluation framework includes direct and indirect probes to both test memorization of updated facts and reasoning over them, for any update learning methods. Next, we present a lightweight method called \textbf{memory conditioned training} (\mrl), which conditions tokens in the update corpus on self-generated ``memory'' tokens during training. Our strategy encourages LLMs to surface and reason over newly memorized knowledge at inference. Our results on two strong LLMs show that (1) \method~ benchmark is highly challenging, with the best CPT models achieving $<\!2\%$ in indirect probing setting (reasoning) and (2) \mrl~training significantly outperforms prior continued pre-training (CPT) baselines, improving direct probing (memorization) results by up to $25.4\%$.
\end{abstract}

\section{Introduction}

Parametric knowledge of large language models (LLMs) \cite{brown2020languagemodelsfewshotlearners} remains mostly static \cite{gekhman2024does} after the pre-training stage, whereas knowledge in the world continues to change. Even within the pre-training data, knowledge from recent years can conflict earlier knowledge. But, the auto-regressive training objective biases LLMs toward surfacing more frequent but not necessarily recent knowledge \cite{cheng2024dated, marjanovic-etal-2024-dynamicqa}. Retrieval-augmented generation (RAG) mitigates these issues to some extent \cite{lewis2020retrieval, nakano2021webgpt}, but can be suboptimal \cite{gao2023enabling}.
In this paper, we focus on continued pre-training (CPT) methods that directly update model parameters to memorize updated facts, which they must surface during inference.

\begin{figure}[t]
    \centering
    \includegraphics[scale=0.33, trim=20mm 195mm 70mm 10mm, clip]{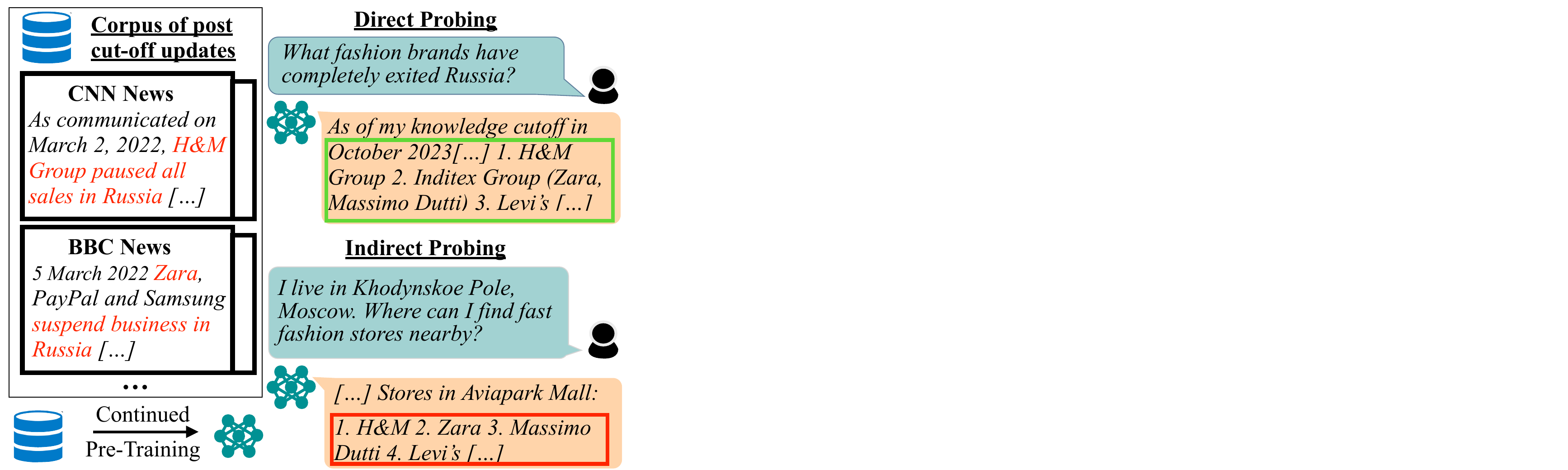} 
    \caption{Example of LLM that is continued pre-trained on updated knowledge surfacing updates in the direct probing but failing under indirect probing settings}
    \label{fig:fig1}
\end{figure}

In our problem setting, a pre-trained LLM with parametric knowledge up to a cut-off date $T$, is continued pre-trained on a corpus of documents reflecting world knowledge updates since $T$. Prior works \cite{ko2024growover, li2024language} explore this problem for a very narrow definition of knowledge update or knowledge conflict, namely the ``entity-substitution'' framework (e.g. \textit{X won \st{two} three awards}). However, updates in the real world are broader and reflect much richer phenomenon (see Figure~\ref{fig:fig1}). They often lead to more nuanced and complex inference-time errors, which entity-substitution framework cannot simulate.

To address these limitations, our paper introduces a new task framework, dataset, and training methodology to adapt LLMs' parametric knowledge to new corpora. First, we introduce \textbf{Knowledge Update Playground} (\method), a framework to automatically instantiate a training corpus with realistic but \textit{fictitious} news articles reflecting knowledge updates. In contrast to prior work that collates recent real updates \cite{li2024language}, we create \textit{fictitious} updates to construct a stable dataset that can be used to study future open-source LLMs with later cut-off dates. Our KUP dataset consists of an evidence corpora of $\sim$55M tokens, which includes news articles and other evidence documents reflecting knowledge updates for 1000 distinct entities. 

KUP's evaluation framework is designed to test both memorization and reasoning capabilities. Consider the example in Figure~\ref{fig:fig1}; an LLM might memorize that H\&M exited Russia, yet still erroneously recommend shopping from H\&M in Moscow when probed indirectly. Ideally, LLMs should identify these conflicts in parametric knowledge and provide temporally consistent responses. To benchmark this for different learning methods, we release a test set of 6260 questions, consisting of \emph{direct probing questions} that test LLMs' memorization of updated knowledge and \emph{indirect probing questions} that require more complex deductive reasoning over these updates.

Next, we introduce a new learning approach memory conditioned training (\mrl) to improve LLM performance on this task. During training, \mrl~prepends ``memory'' tokens, i.e. completions sampled from the model itself and conditioned on a specific entity, to training data about that entity. These completions reflect the base model's parametric knowledge about that entity and encourages LLMs to associate updated knowledge with old memory. We show that LLMs trained using \mrl~are better at surfacing newly learned knowledge at inference compared to baseline training methods. 

We conduct experiments using two strong open source LLMs, LLama-3.1-8B \cite{dubey2024llama} and Mistral-7B-v0.3 \cite{jiang2023mistral}, as the base models. Our results show that \method~is challenging for strong CPT baselines \cite{ko2024growover}, including those with data rephrasing \cite{pieler2024rephrasingnaturaltextdata, maini2024rephrasing}. In fact, all CPT approaches we benchmark report a performance gap of $\sim$30\% compared to an oracle retrieval-augmented upper bound. Surprisingly, continue pre-trained LLMs are better at memorizing high-level  (e.g. triggers, impacts) than low-level details (e.g. where, who).  

Finally, we show that our proposed training method \mrl~outperforms all baselines, improving direct probing results by up to $25.4\%$. Interestingly, we observe that \mrl~can better leverage chain-of-thought (CoT) \cite{wei2023chainofthoughtpromptingelicitsreasoning} at inference, likely because of the parallels between CoT traces at test time and ``memory'' tokens during training. However, our results show that even the best learning methods catastrophically fails in the indirect probing setting, reporting $<2\%$ accuracy for all CPT approaches. This shows that \method~is a challenging test bed for future work to build and improve on. We open source both the codebase and dataset\footnote{Codebase and dataset at: \href{https://huggingface.co/datasets/aochongoliverli/KUP}{https://github.com/Aochong-Li/KnowledgeUpdatePlayground}} to facilitate this. 

To summarize, our key contributions are:
\begin{enumerate}[leftmargin=*, noitemsep,topsep=0pt]
    \item Knowledge Update Playground (\method), a pipeline to automatically instantiate a training corpus of realistic knowledge updates and an evaluation framework to test LLMs' memorization and reasoning  over updates.
    \item Memory Conditioned Training (\mrl), a lightweight continued pre-training method to improve learning of knowledge updates.
    \item Extensive experiments and analysis to benchmark current CPT methods on KUP that surface their key shortcomings. We show that while our proposed approach \mrl~is superior to CPT baselines, all methods primarily learn to memorize updates and fail to reason over their implications. 
\end{enumerate}

\begin{figure*}[t]
    \centering
    \includegraphics[scale=0.23, trim=5mm 180mm 0mm 10mm, clip]{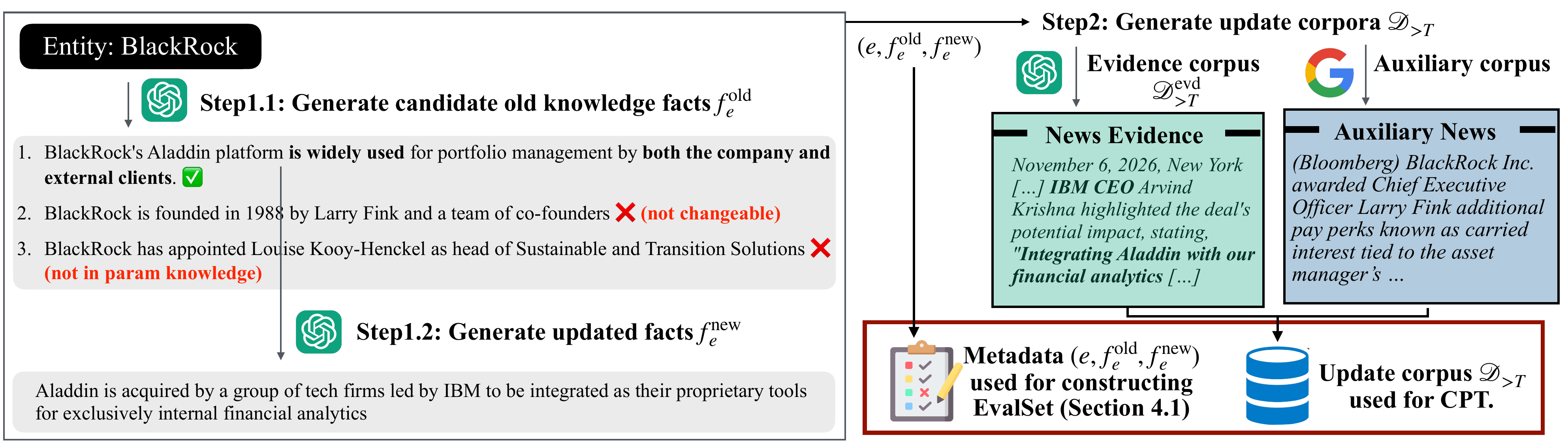}
    \caption{Our \method~data curation pipeline. We omit details about verification of $f^{\mathrm{old}}_e$ (in test models' parametric knowledge?) and $f^{\mathrm{new}}_e$ (contradictory to models' parametric knowledge?). The final training dataset comprises of fictitious evidence documents and real axillary documents for our entity set.}
    \label{fig:pipeline}
\end{figure*}

\section{Knowledge Update Playground (\method)}
\label{sec:pipeline}
First, we describe the \method~framework for automatically create a training corpus that imitates a realistic text corpus reflecting knowledge updates.

\paragraph{Issues with Existing Benchmarks} 
Prior knowledge update benchmarks \cite{ko2024growover, li2024language, marjanovic-etal-2024-dynamicqa} primarily capture and evaluate for entity-substitution phenomena, i.e. swapping certain entities (e.g., names, numbers, locations) in factual statements to simulate conflicts or updates. Although simple and tractable, this framework fails to capture the breadth of real-world knowledge dynamics (see Figure~\ref{fig:fig1}). Moreover, these datasets often update unchangeable facts.\footnote{For example, \textit{Elizabeth II is married to Prince Philip, Duke of Edinburgh} is changed to \textit{Elizabeth II is married to Harry Garnett from 4 September, 2034 to 5 December, 2044} despite both individuals being deceased in \citet{su2024conflictbank}.} In some cases \cite{su2024conflictbank}, the entity-substitution framework is designed to test knowledge editing, i.e. \textit{overwriting} old knowledge with new in model parameters. In the knowledge update setting that we want to study, both old and new knowledge should be retained, but the latter should be surfaced when explicitly or implicitly probed for latest information. 

\paragraph{Notation}
Let $M_{T}$ denote a language model pre-trained on knowledge up to time $T$. $\mathcal{D}_{>T}$ is a corpus of news articles that reflect updates to world knowledge after $T$. The task is to train $M_{T}$ on $\mathcal{D}_{>T}$, resulting in model $M_{{>T}}$. \method~is designed to evaluate how well $M_{{>T}}$ memorizes and reasons over knowledge updates in $\mathcal{D}_{>T}$. It consists of the following components:
\begin{enumerate}[leftmargin=*, itemsep=1pt,topsep=1pt]
    \item \textbf{Knowledge update pairs} $(f^{\mathrm{old}}_e, f^{\mathrm{new}}_e)$: For an entity $e$, $f^{\mathrm{old}}_e$ is an old fact stored in $M_T$'s parameters, whereas $f^{\mathrm{new}}_e$ is a new fact that updates (or contradicts) $f^{\mathrm{old}}_e$. Given a test model, we verify that $M_T$ recognizes $f^{\mathrm{old}}_e$ but not $f^{\mathrm{new}}_e$.
    
    \item \textbf{Evidence document corpus} $\mathcal{D}_{>T}^{\mathrm{evd}}$: consisting of fictitious news articles that reflect the knowledge updates above. We also include an auxiliary collection of real-world news articles for all entities. This helps us emulate a realistic corpus where multiple updates happen to the same entity. The training corpus $\mathcal{D}_{>T}$ combines $\mathcal{D}_{>T}^{\mathrm{evd}}$ and the auxiliary collection.
    
    \item \textbf{Evaluation toolkit} \evaltool: evaluates memorization and reasoning capabilities of $M_{>T}$ over the above updates. We describe this in Section~\ref{sec:evaluation}.
\end{enumerate}

\subsection{Curating \method's Training Corpora}
\label{sec:curateKUP}
In this section, we describe our pipeline for synthesizing $(f^{\mathrm{old}}_e, f^{\mathrm{new}}_e)$ and $\mathcal{D}_{>T}^{\mathrm{evd}}$ (see Figure~\ref{fig:pipeline}).

\paragraph{Step 0: Identify Candidate Entities}
\label{sec:kup_entities}
We identify 10 broad entity categories, e.g. people, companies, landmarks, etc.\footnote{Additional: infrastructure, institutions, sports, technologies, media series, law \& policies, events. } Our desiderata for these entities is (i) \textit{changeability}, which excludes, for e.g., historic events (ii) \textit{reasonable popularity}, to ensure that knowledge about these is present in the parameters of LLMs we use for experiments. For each category, we bootstrap candidate entities by iteratively prompting \textsc{GPT-4o}, starting with a hand-selected seed example set. We include constraint (i) in this prompt itself. To ensure (ii), we generate Wikipedia-style articles using our test LLMs $M_T$ for each candidate entity, and only retain those that report content high overlap with real Wikipedia articles. Appendix~\ref{app:prompt_dataset} outlines this process. We create an entity set $E$ of 1000 entities in this step.

\paragraph{Step 1: Generate Knowledge Update} For each entity $e$, we next generate the $(f^{\mathrm{old}}_e, f^{\mathrm{new}}_e)$ pairs, along with a textual description of an event sequence that can realize this knowledge change.  

The first stage (step 1.1) is to \textbf{collect candidates $\bm{f^{\mathrm{old}}_e}$}, i.e. current facts about entity $e$ that can be changed. We have three broad requirements for these facts: (i) mutable, to filter candidates like \textit{``Geoffrey Hinton invented Boltzmann machines''} or \textit{``Queen Elizabeth II died in 2022''} (ii) plausibly changeable, to avoid stable facts like \textit{``White House is in Washington D.C.,''} and (iii) objective, to filter subjective statements like \textit{``NVIDIA is a visionary AI company.''} Table \ref{tab:additional_examples} gives examples of facts that satisfy these criteria. For each $e$, we generate five such candidates. Concretely, we use a strong LLM (\textsc{GPT-4o}) to propose and filter $f^{\mathrm{old}}_e$. Appendix~\ref{app:prompt_dataset} provides details about prompts and quality control mechanisms.

In a second stage (step 1.2), we \textbf{generate updated facts $\bm{f^{\mathrm{new}}_e}$} for each $f^{\mathrm{old}}_e$ from the previous step. Our aim is to generate updates that are \textit{realistic} and \textit{contradictory} to the prior fact $f^{\mathrm{old}}_e$. We find that strong LLMs like \textsc{GPT-4o} perform better at proposing realistic and logical updates when prompted to additionally generate fictitious event sequences that realize the update from $f^{\mathrm{old}}_e$ to $f^{\mathrm{new}}_e$.

\paragraph{Verifying $\bm{f^{\mathrm{old}}_e}$ and $\bm{f^{\mathrm{new}}_e}$} In order to align with the goals of \method, we need to ensure that LLM $M_T$'s parametric knowledge includes  $f^{\mathrm{old}}_e$ and contradicts $f^{\mathrm{new}}_e$. We probe both our test LLMs to guarantee this. Concretely, we generate answers to True/False questions using test models $M_T$ for both $f^{\mathrm{old}}_e$ and $f^{\mathrm{new}}_e$ facts. We only retain pairs where it generates the expected label (True for $f^{\mathrm{old}}_e$ and False for $f^{\mathrm{new}}_e$) for both. This process filters out roughly 30\% tuples. We retain one $(f^{\mathrm{old}}_e, f^{\mathrm{new}}_e)$ pair per entity after the verification step.\footnote{We found that different $f^{\mathrm{new}}_e$ for the same entity often contradict each other. We avoid this noise in our dataset by retaining only one update per entity.}

\paragraph{Step 2: Generating Training Corpora} The goal of \method~is to simulate a realistic learning environment with continuously evolving knowledge, to develop and evaluate CPT methods. To this end, we need to instantiate grounding evidence documents (e.g. news articles, social media posts, etc.) for each metadata pair $(f^{\mathrm{old}}_e, f^{\mathrm{new}}_e)$. \footnote{Note that many existing benchmarks simply use $f^{\mathrm{new}}_e$ as the grounding evidence \cite{ko2024growover}. However, this is extremely artificial, and any insights (e.g. degree to which LLMs memorize the updated fact statement during CPT) are not guaranteed to transfer to realistic settings.}

We call our grounding news article corpora $\mathcal{D}_{>T}^{\mathrm{evd}}$. To generate $\mathcal{D}_{>T}^{\mathrm{evd}}$, we use the event sequences constructed in the previous step as a guide. We use \textsc{GPT-4o} to generate 5 news articles for each update pair $(f^{\mathrm{old}}_e, f^{\mathrm{new}}_e)$ by conditioning on their event sequence. Appendix~\ref{app:prompt_dataset} provides the prompt details. 

Next, we supplement $\mathcal{D}_{>T}^{\mathrm{evd}}$ with recent news articles on the web for all entities. This ensures that, similar to the real world data, our knowledge update corpus includes multiple, diverse news events per entity, although only knowledge from $\mathcal{D}_{>T}^{\mathrm{evd}}$ is evaluated. In practice, we use SERPHouse\footnote{\url{https://www.serphouse.com/}} API to collect recent news with $e$ as the keyword on \textsc{Google News}. These scrapped news articles constitute the auxiliary data in the corpus $\mathcal{D}_{>T}$.

\paragraph{\method's Continued Pre-Training Setup} As highlighted in red in Figure~\ref{fig:pipeline}, only data from $\mathcal{D}_{>T}$ ($\mathcal{D}_{>T}^{\mathrm{evd}}$ and auxiliary news) from Step 2 is used for continued pre-training; test LLMs cannot directly access the metadata, like the fact statements $(f^{\mathrm{old}}_e, f^{\mathrm{new}}_e)$. We use these later in \S\ref{sec:evaluation} to construct evaluation sets.

\subsection{\method~Dataset Analysis}

\begin{table}[t]
    \centering
    \small
    \begin{tabular}{r|cc} \toprule
        \textbf{Statistic} & \textbf{\# / Entity} & \textbf{\# Total Tokens} \\ \midrule
        \textbf{Fact Updates} & 1 & - \\
        \textbf{Evidence Documents $\bm{\mathcal{D}_{>T}^{\mathrm{evd}}}$} & 5 & 3.3M \\ 
        \textbf{Auxiliary Articles} $\bm{\mathcal{D}_{>T}^{\mathrm{aux}}}$ & 47.6 & 52.4M\\
        \textbf{Total Articles} & 52.6 & 55.7M \\
        \bottomrule
    \end{tabular}
    \caption{\method~dataset statistics. We report token counts using the LLaMA-3.1-8B tokenizer.}
    \label{tab:data_stats}
\end{table}

Table~\ref{tab:data_stats} shows the overall statistics for \method's training dataset for CPT. Our dataset contains 1000 knowledge updates, reflected by evidence corpus $\mathcal{D}_{>T}^{\mathrm{evd}}$ of 3.3M tokens. Including auxiliary corpus leads to a total of 55.7M tokens in $D_{>T}$.

\paragraph{\method~includes richer knowledge update phenomenon than prior benchmarks beyond simple entity substitutions} Figure~\ref{fig:pipeline} shows examples of $(f^{\mathrm{old}}_e, f^{\mathrm{new}}_e)$ pairs in our constructed dataset. More examples are included in Table~\ref{tab:additional_examples} in Appendix.
We first categorize knowledge update $f^{\mathrm{old}}_e \!\rightarrow\!f^{\mathrm{new}}_e$ into two broad categories: \textbf{attribute/entity substitution} and \textbf{contextual rewrite}. An update is classified as attribute substitution if it can be realized by swapping \textit{isolated} entity-based attributes in $f^{\mathrm{old}}_e$ (e.g. \textit{X lives in \st{Boston} NYC}). On the other hand, a contextual rewrite \textit{globally} edits the entire $f^{\mathrm{old}}_e$ statement, changing the fundamental nature of $f^{\mathrm{old}}_e$ with downstream ramifications. In our evidence corpus $\mathcal{D}_{>T}^{\mathrm{evd}}$, entity substitution and contextual rewrite account for 10.2\% and 89.8\% of the updates respectively. On the other hand, entity-substitution comprise 100\% of the updates in prior datasets \cite{ko-etal-2024-growover, su2024conflictbank, li2024language, marjanovic-etal-2024-dynamicqa}.

\paragraph{Analysis of phenomenon} To further compare the characteristics of \method~and prior datasets, we define three properties of knowledge updates. First, we define \textbf{external trigger event}, which applies when events external to the entity directly or indirectly causes the update $f^{\mathrm{old}}_e \!\rightarrow\!f^{\mathrm{new}}_e$  (e.g., \textit{Due to change in India's foreign policy, $X$ decided...}). Next, we define \textbf{narrative augmentation} that refers to inclusion of substantial details of update process. We expect most tokens in $\mathcal{D}_{>T}^{\mathrm{evd}}$ to be these descriptions. Finally, we define \textbf{downstream impact} to refer to inclusion of details about the impact that update $f^{\mathrm{old}}_e\!\rightarrow\!f^{\mathrm{new}}_e$ has on external states. 

\begin{table}[t]
    \centering
    \small
    \begin{tabular}{r|ccc} \toprule
        \textbf{Category} & \textbf{GO} & \textbf{CB} & \textbf{\method} \\ \midrule
        External trigger event & 6 & 4 & 40 \\
        Narrative Augmentation & 4 & 100$^{*}$ & 100\\
        Downstream Impact & 2 & 2 & 42 \\ \bottomrule
    \end{tabular}
    \caption{Analysis of evidence corpus for \textsc{GrowOVER} (GO), \textsc{ConflictBank} (CB) and \method. All entries are percentages (\%). \method~includes richer knowledge update information for research in CPT methods. $^*$CB evidence includes details only for $f^{\mathrm{new}}_e$, assumed to be noise that invalidates (mostly unchangeable) $f^{\mathrm{old}}_e$. It is not designed to study knowledge updating.}
    \label{tab:data_phenom}
\end{table}

We manually annotate 50 data points randomly selected from \textsc{GrowOVER} \cite{ko-etal-2024-growover}, \textsc{ConflictBank}\footnote{By design, 2/3 of the knowledge conflicts in ConflictBank change immutable facts about entities. Since our focus is on mutable facts, we annotate the temporal conflicts subset.} \cite{su2024conflictbank} and our dataset. Table~\ref{tab:data_phenom} shows the distribution. We see that \method~dataset contains richer information about the update $f^{\mathrm{old}}_e \!\rightarrow\! f^{\mathrm{new}}_e$ across all dimensions; $\mathcal{D}_{>T}^{\mathrm{evd}}$ includes details about \textbf{external trigger events and downstream impacts for 40\% and 42\% of \method's updates respectively}. In contrast, \textsc{GrowOVER} dataset, which is also designed for CPT research, is extremely artificial and structurally homogeneous, as less than 6\% of its data includes these properties. 

Although \textsc{ConflictBank} contains evidence documents detailing narrative augmentations for each entity, they primarily focus on $f^{\mathrm{new}}_e$ instead of the update process of $f^{\mathrm{old}}_e\!\rightarrow\!f^{\mathrm{new}}_e$. More concerningly, $f^{\mathrm{new}}_e$ changes immutable facts  $f^{\mathrm{old}}_e$ for most entities.\footnote{Our analysis showed that $>40\%$ of facts in ConflictBank-Temporal were logically impossible, e.g. X getting an award after they have deceased.} While these are reasonable for the task \textsc{ConflictBank} is designed to study, i.e. LLM robustness under noisy conflicts like misinformation, it makes their dataset unsuitable to study continued pre-training under \textit{realistic} knowledge updates. 

\section{Memory Conditioned Training (\mrl)}
Next, we describe a lightweight learning method, ``memory conditioning,'' and explain how to apply it during continued pre-training (CPT) and at inference time for knowledge update corpora.

\begin{figure}[t]
    \centering
    \includegraphics[scale=0.6, trim=85mm 220mm 40mm 30mm, clip]{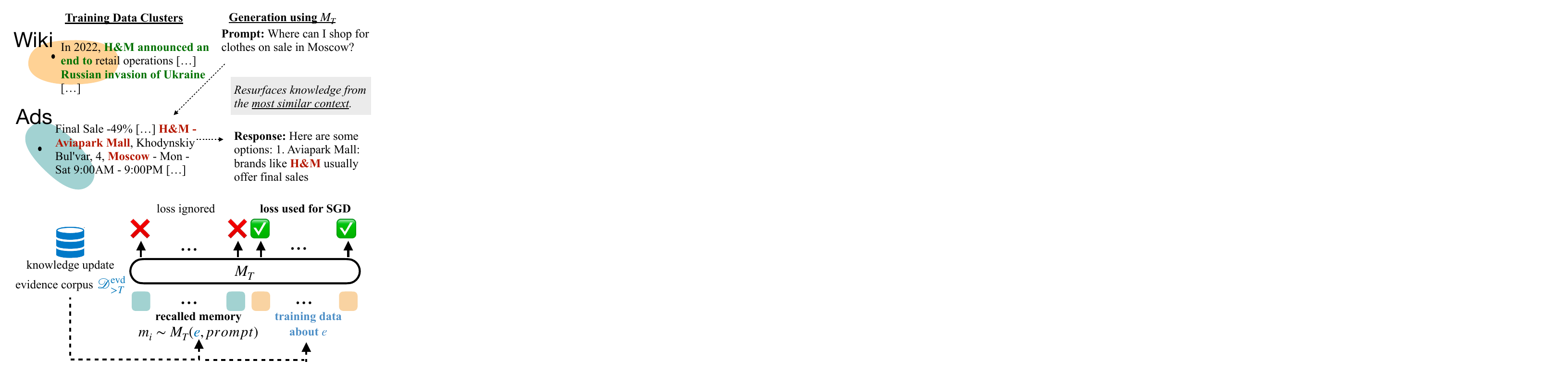}
    \caption{Illustration of Memory Conditioned Training}
    \label{fig:method}
\end{figure}

\paragraph{Motivation} In the pre-training stage of $M_T$, the entity ``H\&M'' is presumably seen with other information in various contexts. In a simplified setting (see the top part of Fig~\ref{fig:method}), assume only two context clusters exist in $M_T$'s memory about ``H\&M'': Wikipedia (Wiki) and Advertisements (Ads), and they present a knowledge conflict. In the former case, ``H\&M'' is contextualized with historical backgrounds (e.g., ``\textit{Russian invasion of Ukraine in 2022}"), whereas in the latter case with store locations (e.g,. ``\textit{Aviapark Mall}"). $M_T$ is likely to memorize both these instances of knowledge during training, but not \textit{resolve} this internal conflict \cite{marjanovic-etal-2024-dynamicqa}. At test time, when prompted with a prefix from Wiki (or Ads), we expect model completions \textbf{to surface knowledge} (e.g., ``end operations" or ``final sales'') \textbf{from the more related context Wiki (or Ads)}. Drawing analogy with our task framework, we hypothesize $M_{>T}$  has memorized $f^{\mathrm{new}}_e$ but still surfaces $f^{\mathrm{old}}_e$ when the prompt is closer to the pre-training distribution $\mathcal{D}_{<T}$. Our training method, \textbf{memory conditioned training (MCT)}, synthetically forces models to learn $f^{\mathrm{new}}_e$ conditioned on prefixes from parametric knowledge about $e$.

\paragraph{Memory Conditioned Training (MCT)} 
As shown in the lower part of Figure \ref{fig:method}, we aim to contextualize training data from $\mathcal{D}_{>T}^{\mathrm{evd}}$, reflecting update $f^{\mathrm{old}}_e \rightarrow f^{\mathrm{new}}_e$, with the parametric memory of $M_T$. However, pre-training datasets for most models are not publicly released. Therefore, we instead sample completions from $M_T$ itself as a proxy for memories about each entity.

In practice, we first prompt $M_T$ to generate a Wikipedia-style output as parametric memory for each entity $e$. During training, we divide the memory into smaller ``memory token'' chunks $m_i$, each covering different memory pieces about $e$, and prepend a random $m_i$ to the training data. This ensures $M_T$ can attend to related ``memory'' elicited from itself when learning knowledge from $\mathcal{D}_{>T}^{\mathrm{evd}}$, and perhaps some $m_i$ may also be associated with $f_e^{\mathrm{old}}$ in the pre-training corpus.

We also modify the language modeling objective to be $- \sum_{n=|m_i|}^{N} \log P_{M_T}(x_n|x_{1:n-1})$ by excluding the loss of $m_i$ tokens from the input sequence $x_{1:N}$  (see Figure \ref{fig:method}).\footnote{Note that the loss from $m_i$ tokens are expected to be low as these already have high probability under $M_T$.} This loss design is to prioritize the training signals from knowledge update rather than to reinforce already acquired knowledge. This design also avoids learning potential conflicting knowledge (i.e. $f^{\mathrm{old}}_e$ from $m_i$ and $f^{\mathrm{new}}_e$  from $\mathcal{D}_{>T}$) in the same training block.

\paragraph{Memory Recall at Inference} At test time, $M_{>T}$ is given a question, and the correct response should reflect knowledge about $f^{\mathrm{new}}_e$. To align with MCT approach, we first construct a prompt to instruct $M_{>T}$ to generate a piece of related ``memory'' for the question, i.e. $memory \sim M_{>T}(prompt)$, and then generate outputs conditioned on both: $response \sim M_{>T}(memory, question)$.

We expect memory ``recalled'' by $M_{>T}$ to reflect its newly acquired knowledge about $e$, helping $M_{>T}$ respond with $f^{\mathrm{new}}_e$. Functionally, $m$ can be viewed as chain-of-thought (CoT) traces \footnote{In later evaluation, we just use ``CoT'' to refer to ``memory recall'' at inference, given that they are functionally equivalent} \cite{wei2023chainofthoughtpromptingelicitsreasoning}, so we also apply it to other baseline methods during inference (\S \ref{sec:results}).

\section{Experiments}
\label{sec:experiment}

We use Llama-3.1-8B (\textsc{LLaMA}; \citet{dubey2024llama}) and Mistral-7B-v0.3 (\textsc{Mistral}; \citet{jiang2023mistral}) for all experiments. We train each model for 1 epoch on $\mathcal{D}_{>T}$. We follow the recipe in \citet{yang2024synthetic} and include 1\% replay data from \textsc{RedPajama} \cite{weber2025redpajama} for all models and baselines (described below). We use a learning rate of 1e-05 for all our experiments. We run both training and inference on 2xH100s. 

\paragraph{Baselines}
We compare our training method MCT against the following baselines: (1) Standard \textbf{CPT} that directly trains $M_{T}$ on the new corpus $\mathcal{D}_{>T}$. (2) CPT with \textbf{Rephrase}, that augments the data in new corpus $\mathcal{D}_{>T}$ by re-writing its contents on different styles (e.g., Reddit posts, Podcast transcripts) \cite{maini2024rephrasing, yang2024synthetic}. We follow \citet{yang2024synthetic} and use a strong LLM (\textsc{GPT-4o}) to curate these data augmentations for each synthesized evidence news article in $\mathcal{D}_{>T}$. 

Note that many update pairs $(f^{\mathrm{old}}_e, f^{\mathrm{new}}_e)$ in \method~ are not and cannot be as structured as entity-substituted changes; prior knowledge editing methods \cite{mengmass, mitchellfast} are not straightforwardly applicable. Moreover, recent work \cite{padmanabhan2024propagating} has shown that standard CPT outperforms these more specialized methods.

\subsection{\evaltool~: Evaluation Protocol}
\label{sec:evaluation}

We evaluate the trained models under two test scenarios: (i) \textbf{direct probing}, and (ii) \textbf{indirect probing} (see Fig~\ref{fig:fig1} and Table~\ref{tab:indirect_probing_example} for examples). 

\subsubsection{Direct Probing}
\label{sec:directprobing}
The goal here is to directly probe if $M_{T}$ \textbf{memorizes} and can \textbf{retrieve} the correct update described in $\mathcal{D}_{>T}$. Therefore, we design two types of questions (multiple choice, free form) to study this: 

\paragraph{Multiple-choice questions (MCQ)} We test the ability of $M_{>T}$ to identify the updated knowledge $f_e^{\mathrm{new}}$ among four options about $e$ in a multiple choice setting. We build two MCQ tests: (i) \textbf{update vs. distractors} with the goal of selecting the gold $f_e^{\mathrm{new}}$ among three misleading options, and (ii) \textbf{update vs. old} that asks to select $f_e^{\mathrm{new}}$ over two misleading options and $f_e^{\mathrm{old}}$. Each MCQ test contains 1K questions, one for each entity. Appendix \ref{app:app_eval_prompts} details how the distractors are generated. Note that \method's pipeline automatically generates statements $f_e^{\mathrm{new}}$ and $f_e^{\mathrm{old}}$ as metadata (Step 1 in \S\ref{sec:pipeline}), which we directly use to construct the tests.

\paragraph{Free-form questions} We further probe which evidence details from $\mathcal{D}_{>T}^{\mathrm{evd}}$ are learned beyond the update statements $f_e^{\mathrm{new}}$. We generate roughly 4 questions per update, asking for the trigger event, downstream impact, or other more granular details about the update.\footnote{We use GPT-4o to generate these. Table \ref{tab:freeform_qa_prompt} includes the prompt for question generation and quality control.} Overall, we generate 4.2K questions, including 1.1K questions about triggers and impact, and 3.1K questions about update event details. Table \ref{tab:freeform_qa_example} shows examples of each At test time, we provide the update statements $f_e^{\mathrm{new}}$ to $M_{>T}$ for context along with the question, and generate free-form responses. We use \textsc{GPT-4o-mini} as our LLM judge for evaluation, comparing model responses against the evidence articles. 

\begin{table}[t]
    \centering
    \small
    \begin{tabularx}{\columnwidth}{@{}X@{}}
        \textbf{\textsc{Entity}}: Volvo XC40 Recharge\\
        \midrule
        \textbf{Old}: Volvo offers the XC40 Recharge with over-the-air updates for its software and infotainment system.\\
        \textbf{Update}: Volvo's XC40 Recharge's software and infotainment systems are switched to a proprietary Volvo OS.\\
        \midrule
        \textbf{Event Details Question}: \textbf{When} did Volvo begin using its proprietary Volvo OS for vehicle software platforms?\\
        \midrule
        \textbf{Trigger \& Impact Question}:  \textbf{Why} does Volvo prefer USB drive updates over over-the-air updates?" \\
    \end{tabularx}
    \caption{Examples of event detail and trigger \& impact free-form questions for the same knowledge update}
    \label{tab:freeform_qa_example}
\end{table}

\begin{table}[t]
    \centering
    \small
    \begin{tabularx}{\columnwidth}{@{}X@{}}
        \textbf{\textsc{Entity}}: Edinburgh International Science Festival\\
        \textbf{\textcolor{red}{\textsc{Old}}}: The festival receives ... \textcolor{red}{private sponsorships}.\\
        \textbf{\textcolor{blue}{\textsc{Update}}}: ... {eliminating all private sponsorships}.\\
        \midrule
        \textbf{\textsc{Question}}: What events does Baillie Gifford  still sponsor?\\
        \midrule
        \textbf{\textsc{Model} \textsc{Response}}:  Baillie Gifford \& Co. may still sponsor: 1. \textcolor{red}{Edinburgh International Science Festival} 2. ..." \\
        \midrule
        \textbf{\textsc{Entailment}}: \textcolor{red}{Old Knowledge}
    \end{tabularx}
    \caption{A failure case under indirect probing. Red: old fact statement $f_e^{\mathrm{old}}$; blue: updated fact statement $f_e^{\mathrm{new}}$}
    \label{tab:indirect_probing_example}
\end{table}

\subsubsection{Indirect Probing} 

Finally, we create a test set to evaluate $M_{>T}$'s \textbf{reasoning} ability to deduce from updated knowledge and apply for indirect probing questions (see Figure~\ref{fig:fig1} for an example). To create these questions, we fix the format of indirect probes to be questions asking for list-style responses. We call this setting indirect probing because the questions do not explicitly mention $e$, but are designed such that models are likely to include information about $e$ in their response (see the example in Table~\ref{tab:indirect_probing_example}). We evaluate whether the response correctly excludes any knowledge from its pre-training, such as $f^{\mathrm{old}}_e$, that conflicts with updated knowledge $f^{\mathrm{new}}_e$.

We find that strong LLMs like \textsc{GPT-4o} cannot be reliably prompted to construct such questions, so we manually curate a small test set of 60 questions. We choose entities for which both the Llama and Mistral test models correctly answer the MCQs from \S\ref{sec:directprobing}. This allows us to evaluate whether LLMs can \textit{reason} over updates they have already memorized in a targeted manner.

For each test example, we report the fraction of times the trained model generated a response that entails the $f^{\mathrm{old}}_e$ knowledge vs. the $f^{\mathrm{new}}_e$ knowledge. Note that there may exist cases where the model generation does mention entity $e$, and therefore, cannot be classified as entailing either. We report the fraction of such  cases as ``N/A'' in Table \ref{tab:indirect_probing_results}. 

\section{Results}
\label{sec:results}

\subsection{Direct Probing}
\label{sec:directprobingresults}
Table~\ref{tab:mcq_performance} outlines the test results (update vs. distractors, update vs. old) for our ``memory conditioned learning'' (\mrl) and all CPT baselines in the \textbf{direct probing w/ MCQs} evaluation setting. We report results for 4-shot and 4-shot + CoT settings. Note that \mrl~includes CoT, i.e. recalling memory at inference, by default; therefore, we additionally report performance without CoT.

\begin{table}[t]
    \centering
    \small
    \renewcommand{\arraystretch}{1.1}  
    \setlength{\tabcolsep}{2pt}        
    \begin{tabular}{r|cc|cc}
    \toprule
    \textsc{Method}
    & \multicolumn{2}{c|}{\textbf{\textsc{Update vs. Dist.}}}
    & \multicolumn{2}{c}{\textbf{\textsc{Update vs. Old}}} \\ 
    & \textsc{LLaMA} & \textsc{Mistral} & \textsc{LLaMA} & \textsc{Mistral} \\ 
    \midrule
        \textbf{\textsc{No-train}}  
      & 14.2 & 16.1 & 1.0$_{\text{(93.0)}}$ & 2.9$_{\text{(90.7)}}$ \\
     \textcolor{ForestGreen}{\textsc{ + CoT}}
      & 21.9 & 26.4 & 3.3$_{\text{(87.6)}}$ & 2.7$_{\text{(91.8)}}$ \\ 
    \midrule
    
    \textbf{\textsc{CPT}}  
      & 20.0 & 17.4 & 5.7$_{\text{(83.6)}}$ & 4.3$_{\text{(84.3)}}$ \\
     \textcolor{ForestGreen}{\textsc{ + CoT}}
      & \underline{41.5} & 34.5 & 17.3$_{\text{(67.2)}}$ & 4.2$_{\text{(88.3)}}$ \\ 
    \midrule
    \textbf{\textsc{Rephrase}}
      & 25.7 & 37.8 & 8.5$_{\text{(80.0)}}$ & 28.8$_{\text{(50.8)}}$ \\
    \textcolor{ForestGreen}{\textsc{ + CoT}}
      & 41.1 & \underline{58.9} & \underline{19.7}$_{\text{(66.1)}}$ & \textbf{40.4}$_{\text{(41.4)}}$ \\ 
    \midrule
    \textbf{Ours} \textcolor{Red}{\textsc{- CoT}}
      & 30.1 & 28.4 & 7.6$_{\text{(83.9)}}$ & 8.7$_{\text{(79.3)}}$ \\
    \textbf{\textsc{Ours}}
      & \textbf{60.7} & \textbf{71.0} & \textbf{45.1}$_{\text{(45.4)}}$ & \underline{34.5}$_{\text{(57.8)}}$ \\ 
    \midrule
    \textbf{\textsc{RAG}} (k=5)
      & 93.0 & 93.0 & 82.2$_{\text{(15.9)}}$ & 74.5$_{\text{(20.8)}}$ \\ 
    \bottomrule
    \end{tabular}

    \caption{Model Performance on Update vs. Dist. and Update vs. Old MCQ w/ \& w.o. CoT. We use boldface and underline to represent best and 2nd best CPT performance. \% times $f_{e}^{\text{old}}$ is chosen in the Update vs. Old setting is reported in parenthesis. We find that old knowledge is preferred over updated knowledge for all CPT approaches. \mrl~shows the largest improvement when using CoT, outperforming baselines in 3/4 settings. } 
    \label{tab:mcq_performance}
\end{table}

\paragraph{LLMs after continued pre-training (CPT) can select the correct knowledge update over misleading choices but still prefer the old knowledge} We find that all $M_{>T}$ models with CoT perform better than the No-Train baseline (i.e. base model without CPT) in the update vs. distractors setting.\footnote{The No-train model performance is below random guess (25\%) because we constructed our distractors options to be very misleading, and in many cases, more likely than the correct update $f^{\mathrm{new}}$, given the pre-trained knowledge of $M_T$.}  However, across the board, models prefer the old knowledge $f^{\mathrm{old}}_e$ (\% times chosen shown in subscript) over the updated knowledge $f^{\mathrm{new}}_e$ in $\mathcal{D}_{>T}^{\mathrm{evd}}$. 

Although our focus is on CPT, we also report results using a simple RAG framework to provide an upper bound. We note that our task, by design, is trivial for RAG.\footnote{The retrieval query, i.e. question with four choices, contain verbatim text from the document corpus. Given correct passages, direct probing MCQs are not hard for strong LLMs.} We divide the update corpus $\mathcal{D}_{>T}$ into chunks of 256 tokens, and use the question and the four choices as the query for retrieval. We use NV-Embed-v2 \cite{lee2025nvembedimprovedtechniquestraining} as the embedding and retrieval model.

\paragraph{Memory conditioned training (MCT) outperforms continued pre-training (CPT) baselines.} It outperforms baselines by 12.1\% to 53.6\% in update vs. distractors setting. In update vs. old setting, our approach improves performance from 25.4\% to 39.4\% for Llama-8B model and performs second best, behind Rephrase, for Mistral-7B. In general, we find that Mistral-7B exhibits different performance improvement behaviors than Llama-8B.

\paragraph{Surprisingly, CoT improves performance in our knowledge-intensive  task} This results sits in contradiction with findings from recent works \cite{sprague2024cot} that CoT primarily helps with reasoning tasks. We hypothesize that although models store the updated knowledge in their parameters, they cannot resolve internal memory conflicts from keeping copies of both old and updated knowledge. CoT or ``recalling memory'' helps as LLMs are better at reasoning over contextual than their internal knowledge.  Also, we observe that the biggest improvement in both update vs. distractors (28.4\% $\rightarrow$ 71.0\%) and update vs. prior (7.6\% $\rightarrow$ 45.1\%) is observed for our MCT training method. We hypothesize that this is because MCT, which appends parametric model-generated related ``memory'' tokens before updated knowledge data, more aligns with the CoT procedure at inference.

\paragraph{\mrl~also outperforms the best performing baseline in free-form QA settings} Table~\ref{tab:freeform_qa} outlines our results for free-form QA; we compare against the best-performing baseline \textit{Rephrase + CoT} from Table~\ref{tab:mcq_performance}. We also observe that \mrl~consistently outperforms this baseline for both question types (``\textit{Trigger \& Impact}'', ``\textit{Event Details}'') and for both Llama-8B and Mistral-7B. Interestingly, we find that \textbf{both models are better at answering more abstract, plot-related questions} (e.g., causes, impacts) \textbf{than low-level details} of the event (e.g., person names, numbers, etc.).

\begin{table}
    \centering
    \small
    \renewcommand{\arraystretch}{0.9}
    \begin{tabularx}{\columnwidth}{>{\centering\arraybackslash}p{1.3cm} | >{\centering\arraybackslash}p{1.1cm} | >{\centering\arraybackslash}X | >{\centering\arraybackslash}X | >{\centering\arraybackslash}X} \toprule
        Method & \multicolumn{2}{c|}{Trigger \& Impact} & \multicolumn{2}{c}{Event Details} \\ 
        \midrule
        & \textsc{LLaMA} & \textsc{Mistral} & \textsc{LLaMA} & \textsc{Mistral} \\
        \midrule
        Re + CoT  & 56.8 & 61.1 & 34.9 & 47.0 \\
        Ours  & \textbf{65.9} & \textbf{68.6} & \textbf{52.5} &  \textbf{64.3} \\
     \bottomrule
    \end{tabularx}
    \caption{accuracy (\%) for 1) trigger \& impacts, 2) event details questions; ``Re'': CPT w/ data rephrasing}
    \label{tab:freeform_qa}
\end{table}

\subsection{Indirect Probing}
\label{sec:indirectprobing_results}

\begin{table}[t]
    \centering
    \small                      
    \renewcommand{\arraystretch}{1.0} 
    \setlength{\tabcolsep}{10pt}      
    \begin{tabularx}{\columnwidth}{>{\centering\arraybackslash}p{1.2cm} | >{\centering\arraybackslash}p{1.5cm} | >{\centering\arraybackslash}X | >{\centering\arraybackslash}X | >{\centering\arraybackslash}X}
        \toprule
        \textbf{\textsc{Model}} & \textbf{\textsc{Method}} & \textbf{\textsc{Upd.}} & \textbf{\textsc{Old}} & \textbf{\textsc{N/A}} \\
        \midrule
        \multirow{3}{*}{\makecell{\textbf{\textsc{LLaMA}}}} 
        &CPT + CoT& 0.5 & 84.8 & 14.7 \\
        &Re + CoT& 0.3 & 77.7 & 22.0 \\
        &Ours& 0.8 & 83.2 & 16.0 \\
        
        \midrule
        \multirow{3}{*}{\makecell{\textbf{\textsc{Mistral}}}} 
        &CPT + CoT& 1.1 & 80.2 & 18.7 \\
        &Re + CoT& 3.0 & 80.3 & 16.7 \\
        &Ours& 1.8 & 78.5 & 19.7 \\
        \midrule
        \multirow{2}{*}{\makecell{\textbf{\textsc{RAG}} \\ (k=5) }}     & $\text{Retrieved}$ & 16.8 & 66.2 & 17.0 \\
        & $\text{Oracle}$ & 69.6 & 25.5 & 4.8 \\
        \bottomrule
    \end{tabularx}
    \caption{Percentage (\%) of model answer entailment for indirect probings. \textsc{UPD.}: updated knowledge; \textsc{Old}: old knowledge; \textsc{N/A}: no entailment. ``Re'': CPT w/ data rephrasing; ``retrieved'': retrieved passages; ``oracle'': ground-truth passages for updated knowledge. }
    \label{tab:indirect_probing_results}
\end{table}

For these experiments, we first train $M_T$ using the proposed MCT method, i.e. the best performing training method in \S \ref{sec:directprobingresults}. Since indirect probing involves question answering, we then supervised fine-tune $M_{>T}$ (i.e. LLMs after training) on the set of 4.2K Q\&A pairs generated for free-form direct probing in \S \ref{sec:directprobing}. We also report RAG results with retrieved and oracle passages. Table \ref{tab:indirect_probing_results} outlines the results for indirect probing for the instruction-tuned models with CoT and RAG.

\paragraph{All continued pre-trained (CPT) LLMs fail catastrophically at indirect probing} We found that the model responses entail $f^{\mathrm{old}}_e$ for 78.5\% to 83.2\% of the cases. On the other hand, the trained model only avoid errors by surfacing the updated knowledge ($f^{\mathrm{new}}_e$) or neither updated nor prior knowledge less than a combined 20\% of cases. 

Unlike direct probing, we find that RAG does not straightforwardly solve indirect probing questions. Even with oracle passages in the context, LLMs still entail old knowledge 25\% of the times.\footnote{We note that the RAG performances can potentially be improved by iteratively retrieving and self-correcting generated text. However, we omit those experiments as RAG is not the focus of this work.}

\section{Analysis}
\label{sec:analysis}
\begin{table}[t]
    \centering
    \small
    \begin{tabularx}{\columnwidth}{@{}X@{}} \toprule
        \textbf{\textsc{CoT memory recall}}: Jamie Joseph ... was suspended indefinitely following a controversial rule violation ... which uncovered that Joseph had \textbf{\textcolor{ForestGreen}{unauthorized communications with players during a critical match}} against \textcolor{red}{\sout{\textbf{Scotland}}} \textcolor{red}{\textbf{(\textcolor{ForestGreen}{South Korea})}} in the \textcolor{red}{\sout{\textbf{2025}}} \textcolor{red}{\textbf{(\textcolor{ForestGreen}{2026})}} \textbf{\textcolor{ForestGreen}{Rugby World Cup}}. Answer: A.\\ \bottomrule
    \end{tabularx}
    \caption{CoT memory recall example. Green: grounded extra context; Red: hallucinated extra context}
    \label{tab:memory_recall_example}
\end{table}

\paragraph{Analysis of CoT for recalling memory} To understand what $M_{>T}$ actually ``recalls'' at test time, we manually annotate 100 such CoT traces, like the example in Table \ref{tab:memory_recall_example} (for both Llama and Mistral) only for MCQs they answer correctly (\S\ref{sec:directprobingresults}). We classify these CoTs into 3 categories based on the content: (i) direct copying/rephrasing content from MCQ prompt without ``recalling'', (ii) containing additional ``recalled'' memory grounded in $\mathcal{D}_{>T}^{\mathrm{evd}}$, and (iii) containing hallucinated ``recalled memory.'' 

We find that: copying MCQ options verbatim (case i) accounts for 25\% of the CoTs. When the CoT contains ``recalled'' memory, we find that, on average, it includes 1.68 \textit{concrete} details pulled from evidence news articles. 58\% of these details are verifiably grounded in $\mathcal{D}_{>T}^{\mathrm{evd}}$ (case ii). At an example level, 35\% of the examples do not have \textit{any} hallucinated details.

\paragraph{Continued pre-training (CPT) perplexity does not suggest how well $M_{>T}$ memorizes updated knowledge} As LLMs are trained to maximize log probability of $D_{>T}$, we ask, ``does lower perplexity align with higher downstream performance on \method?'' We report results in Table \ref{tab:perplxity_updatememorization_mcq} and \ref{tab:perplxity_updateprior_mcq} in the Appendix for direct probing with MCQs. We find i) no difference in perplexity of evidence news articles for correctly and incorrectly answered questions and ii) new knowledge corpus has much lower perplexity than old knowledge, but this does not translate to preferring the former at test time. This shows that simply minimizing auto-regressive loss may not improve model performance on \method~or similar tasks.

\section{Related Works}

\paragraph{Continued Pre-training}Continued pre-training  \cite{gururangan2020dontstoppretrainingadapt} has shown to be a cost-efficient and effective method for adapting language models to new domains \cite{rozière2024codellamaopenfoundation, chen2023meditron70bscalingmedicalpretraining, lu2024mathcoder2bettermathreasoning} and update-to-date knowledge \cite{jin-etal-2022-lifelong, jang2022continualknowledgelearninglanguage, qin2022elleefficientlifelongpretraining}. Prior work \cite{parmar2024reusedontretrainrecipe, guptacontinual, ibrahim2024simplescalablestrategiescontinually, rolnick2019experiencereplaycontinuallearning, chen2024continualmemorizationfactoidslarge} has worked on methods to alleviate alleviate issues like catastrophic forgetting \cite{ROBINS01061995} and improve model's learning ability through data augmentation techniques \cite{yang2024syntheticcontinuedpretraining, ding2024data, zhang2022simple}.We extend this line of work by proposing memory conditioned training (\mrl) for updating model parametric knowledge with new knowledge corpora.

\paragraph{Knowledge Conflict Datasets}To study LLM behaviors in presence of misinformation and outdated information, prior work \cite{longpre-etal-2021-entity, su2024conflictbank, ko2024growover} construct simple, synthetic knowledge conflicts using entity-substitutions. Other lines of work localize and analyze model intra-memory conflicts and effective knowledge cutoffs \cite{marjanovic-etal-2024-dynamicqa, cheng2024dated}. To assess model's ability to provide update-to-date information in the real world, prior benchmarks \cite{vu-etal-2024-freshllms, kasai2024realtimeqawhatsanswer, borkakoty2024chewdatasetchangingevents} also try to collect knowledge updates and evaluate LLMs' behaviors in the real world. Notably, \citet{vu-etal-2024-freshllms} show that RAG frameworks fail for simple queries as real-world data is extremely noisy.

\section{Conclusion}
In this paper, we introduce the Knowledge Update Playground (KUP), a novel framework designed to systemically study the effectiveness of different continued pre-training (CPT) methods for updating large language models' parameters with evolving knowledge. Unlike prior benchmarks that rely on entity-substitution frameworks, KUP curates synthetic datasets that can capture the nuances and complexity of real-world knowledge dynamics. To evaluate LLMs' memorization and reasoning capabilities over knowledge updates, we also develop an evaluation toolkit, KUPEval, which includes both direct and indirect probing tests.

Additionally, we propose Memory Conditioned Training (\mrl), a lightweight yet effective continued pre-training technique. \mrl~outperforms CPT baselines multiple direct probing tests (both MCQ and free-form QA). Nevertheless, indirect probing tasks remain particularly challenging for all existing training methods, and we encourage future research to continue to work on this problem.

\section{Limitations}
Our proposed framework \method~uses \textsc{GPT-4o} to synthetically generate realistic knowledge updates and evidence articles. Due to the cost intensive nature of the task, we restrict the dataset to 1000 knowledge updates. We will release our dataset construction methodology for future works to expand on. Moreover, we conduct our experiments on two LLMs in the 7B-8B scale. Continue pre-training behaviors may differ for larger models with more memorization capacity. We leave this exploration to future work.

\bibliography{custom}

\appendix
\section{Additional Evaluation: retention of prior knowledge after continued pre-training}

\subsection{General Knowledge}
\label{sec:mmlu_appendix}
To ensure that continued pre-training (CPT) does not cause LLMs to catastrophically forget general knowledge from pre-training distribution, we evaluate model on Measuring Massive Multitask Language Understanding (MMLU) \cite{hendryckstest2021} with lm-evaluation-harness \cite{eval-harness} package, see Table \ref{tab:mmlu}. Here we only evaluate LLMs after standard CPT and do not observe significant degradation on knowledge benchmark. We expect CPT with data rephrasing and memory conditioned training (MCT) to have similar results.

\begin{table}[h]
\centering
\begin{tabular}{l|cc}
\toprule
\textbf{Method} & \textbf{LLaMA3-8B} & \textbf{Mistral-7B} \\
\midrule
Base & 65.04 & 62.34 \\
CPT & 64.69 & 60.68 \\
\bottomrule
\end{tabular}
\caption{MMLU scores . Continued pre-training (CPT) does not significantly affect models’ general knowledge as measured by MMLU.}
\label{tab:mmlu}
\end{table}

\subsection{Prior Knowledge of $f_{e}^{\mathrm{old}}$ in KUP}
We conduct an additional experiment to ensure that LLMs still retain prior knowledge (described in $f_{e}^{\mathrm{old}}$) that is updated in KUP. Similar to the verification step in \S \ref{sec:curateKUP}, we ask models to output a True/False label for each $f_{e}^{\mathrm{old}}$ statement in the KUP dataset, and we set the system prompt to ``Today’s Date: December 2023'' so that LLMs should know to use $f_{e}^{\mathrm{old}}$ instead of $f_e^{\mathrm{new}}$. 

The table below shows the percentage of times models correctly output  ``True,'' indicating retention of prior knowledge. All continued pre-trained models choose ``True'' for $\!>\!97\%$ of times for $f_{e}^{\mathrm{old}}$.

\begin{table}[h]
\centering
\begin{tabular}{l|cc}
\toprule
\textbf{Model} & \textbf{LLaMA3-8B} & \textbf{Mistral-7B} \\
\midrule
CPT & 97.7 & 99.7 \\
Rephrase & 99.2 & 99.8 \\
Ours & 99.4 & 97.1 \\
\bottomrule
\end{tabular}
\caption{Percentage (\%) of time LLMs after continued pre-training on knowledge update in KUP still report old knowledge statement $f_{e}^{\mathrm{old}}$ as ``True''}
\label{tab:oldknowledge_cpt}
\end{table}

\section{Examples of Our Dataset}
Here, we include three additional examples from our \method~dataset. None of these knowledge updates are based on or can be achieved by entity-substitution framework. In the example of ``Intuit Inc.,'' updated fact introduces a hypothetical change of QuickBooks merging with Intuit, therefore invalidating the old fact that Intuit owns Mint.
\begin{table*}[th]
    \centering
    \small
    \begin{tabularx}{\textwidth}{@{}lX@{}} 
    \toprule
    \textbf{Entity} & Intuit Inc. \\
    \textbf{Old Fact} & Intuit Inc. owns the personal finance app Mint, which offers budgeting and financial tracking tools. \\
    \textbf{Updated Fact} & Intuit fully integrates Mint’s functions into QuickBooks and Mint ended its operations as a personal finance service. \\
    \textbf{Evidence} & [...] This merger capitalizes on QuickBooks’ burgeoning user base, which reached over 7.5 million by late 2025, presenting Intuit with a substantial opportunity to consolidate its services. The integration enables QuickBooks users to access new personal finance tools, which include detailed spending insights, personalized financial planning tips, and the innovative MintSights™ feature[...] \\
    \midrule
    \textbf{Entity} & Zendaya Coleman \\
    \textbf{Old Fact} & Zendaya Coleman is involved in various fashion projects, working with luxury fashion brands. \\
    \textbf{Updated Fact} & Zendaya Coleman had a fallout with major luxury brands after she was involved in a scandal over controversial fashion ads. This resulted in severance of all professional ties and prohibiting her from future opportunities. \\
    \textbf{Evidence} & [...] the beloved actress and fashion icon Zendaya Coleman has found herself at the center of a public relations storm, severing professional ties with several high-profile luxury brands after a controversial advertisement ignited widespread criticism. The luxury fashion advertisement, which debuted on March 1, quickly became a focal point of contention for its alleged cultural insensitivity, leading to the fallout.[...] \\
    \midrule
    \textbf{Entity} & COP - United Nations Climate Change Conference \\
    \textbf{Old Fact} & The main goal of COP conferences is to assess progress in dealing with climate change and to negotiate commitments from different countries. \\
    \textbf{Updated Fact} & COP conferences are reduced to ceremonial events with no meaningful progress assessment or negotiations, and countries decide on bilateral or regional agreements instead. \\
    \textbf{Evidence} & [...] leading nations unveiled several significant bilateral agreements on the eve of COP 31. The European Union and the United States, for instance, announced a groundbreaking Green Technology Exchange program with an investment of \$50 billion over the next decade. This initiative aims to foster joint innovations in renewable energy through collaborative research, patent sharing, and investment in clean-tech startups, addressing urgent imperatives much faster than the traditional routes of multilateral consensus.[...] \\
    \bottomrule
    \end{tabularx}

\caption{Additional examples of (entity, old fact, updated fact, evidence news article) in \method~dataset}
\label{tab:additional_examples}
\end{table*}

\section{Training Details}
We train all of our models on 2 Nvidia H100s and use the following hyparameters for continued pre-training and supervised fine-tuning (in \S \ref{sec:indirectprobing_results}): learning rate = 1e-5, block size = 2048, batch size = 16, weight decay = 0.01, warm up = 0.05.

\section{Evaluation Details}
\label{app:app_eval_prompts}
In the direct probing setting (\S \ref{sec:directprobing}), we use prompt in Table~\ref{tab:mcq_prompt} to generate misleading options/distractors, which are used in \textit{update vs. distractors} and \textit{update vs. prior} MCQs. The prompt used for generating Q\&A pairs for \textit{free-form questions} is provided in \ref{tab:freeform_qa_prompt}.

\begin{table*}[th]
    \centering
    \ttfamily
    \small
    \begin{tabular}{p{0.9\textwidth}}
    \toprule
Background: You are a helpful research assistant tasked with creating a set of answer choices for a multiple-choice question.\\
Task: You are given an entity, a statement, and a corresponding Wikipedia page. For the provided question, formulate four answer choices:\\
- Choice A: The correct answer, directly supported by the provided statement. Do not include any modifiers (e.g., "strictly," "actively," "entirely," etc.).\\
- Choices B, C, D: Each should be a factually incorrect claim that appears detailed and specific by referencing a unique aspect from the provided Wikipedia. Ensure that each misleading choice includes specific details, even though they are false.\\
\\
Format to Structure Question:\\
A: [Correct choice]\\
B: [Misleading choice]\\
C: [Misleading choice]\\
D: [Misleading choice]\\
\\
Example:\\
Entity: Gigi Hadid\\
Statement: Gigi Hadid chooses to represent herself and manages her own modeling career, entirely separating from agency representation.\\
Question: Which of the following about Gigi Hadid is True?\\
A: Hadid chose to represent and manage her career without agency representation.\\
B: Hadid announced on Instagram her second marriage to Zayn Malik.\\
C: It is revealed that Hadid's foundation didn't donate to Ukrainian victims, leading to online controversies.\\
D: Hadid's clothing line Guest in Residence tried to enter Chinese consumer market\\
\\
Entity: \{entity\}\\
Statement: \{update\}\\
Wikipedia: "\{wiki\}"\\
\\
Requirements:\\
1. All misleading choices (B, C, D) should be longer than the correct choice in length. Misleading choices also should include more details, such as number, names, locations, than the correct choice. But do not introduce any specific date details.\\
2. All choices (A, B, C, D) should strictly use the same time tense and be written in the same sentence structure to be stylistically indistinguishable\\
3. Always refer to the entity by its name rather than using pronouns.\\
4. Do not include additional comments after the question\\
\\
Question: Which of the following about \{entity\} is True?\\
    \bottomrule
    \end{tabular}
    \caption{Prompt for generating Update vs. Distractors and Update vs. Old MCQ}
    \label{tab:mcq_prompt}
\end{table*}

\begin{table*}[th]
    \centering
    \ttfamily
    \small
    \begin{tabular}{p{0.9\textwidth}}
    \toprule
You are a helpful research assistant. Generate a set of 20 to 30 Q\&A pairs from the article below, formatted as a list of JSON objects with ``content'' and ``role'' as keys. ``role'' should be either ``user'' or ``assistant.'' Ensure proper JSON formatting.\\
\\
Template examples of Q\&A pairs:\\
\{template\_qa\}\\
\\
This is the source article:\\
\{article\}\\
\\
Instructions:\\
1. Self-contained questions: Each question must be understandable without requiring the article as context. Each question should include specifics such as names, dates, events, or changes. Avoid anaphoric or vague noun phrases, like ``the person,'' ``the article,'' ``the event,'' ``the transition'' etc. Readers cannot access the article content nor know what transition has happened, so clarify all the references.\\
2. Independent questions: Each question must stand alone and will be presented individually. Do not assume the reader has seen previous questions. Avoid referencing other questions or relying on their background for context. Each question should be fully self-explanatory.\\
3. Diversity of questions: Generate 20 distinct and meaningful questions covering different key aspects of the article.\\
4. Supported answers: Each answer must be correct and grounded in the article, providing supporting evidence or key details.\\
5. Avoiding Quotation Marks: Ensure all double quotes inside JSON values are properly escaped to prevent syntax errors in Python. If quotation marks are necessary within content, use single quotes ('') instead.\\
\\
Additional Instructions:\\
1. Change-oriented question: Given that the article focuses on recent changes, include 1 to 3 simple questions that elicit answers contrasting before and after the change naturally.\\
2. Contextualized answer: For change-oriented questions, ensure answers describe both the previous and updated states of the entity. For example, an answer should explain what was true before the change, when the change occurred, and how the fact evolved into its new state.\\
3. You do not need to differentiate these Q\&A pairs from others. Include all questions in the same list of JSON objects.\\
    \bottomrule
    \end{tabular}
    \caption{Prompt for Free-form QA}
    \label{tab:freeform_qa_prompt}
\end{table*}

\subsection{Perplexity Analysis on Direct Probing w/ MCQ}
As described in \S \ref{sec:analysis}, we measure the perplexity of old knowledge (fact statement $f_{e}^{\mathrm{old}}$) and updated knowledge (evidence articles from $\mathcal{D}_{>T}^{\mathrm{evd}}$) for questions that are answered correctly and incorrectly in direct probing with MCQs. The results from Table \ref{tab:perplxity_updatememorization_mcq} and Table \ref{tab:perplxity_updateprior_mcq} don't show any clear difference in perplexity between correctly and incorrectly answered questions.

\begin{table*}[ht!]
    \centering
    \renewcommand{\arraystretch}{1.2} 
    \setlength{\tabcolsep}{8.5pt}      
    \begin{tabularx}{0.8\textwidth}{>{\raggedright\arraybackslash}p{1.5cm} | 
                                    >{\centering\arraybackslash}p{1.7cm} | 
                                    >{\centering\arraybackslash}X
                                    @{\hskip 9pt} 
                                    >{\centering\arraybackslash}X 
                                    @{\extracolsep{1pt}} | 
                                    >{\centering\arraybackslash}X 
                                    @{\hskip 9pt} 
                                    >{\centering\arraybackslash}X}
        \midrule
        \textbf{\textsc{Model}} & \textbf{\textsc{Training}} 
        & \multicolumn{2}{c|}{\textbf{\textsc{Old Perplexity}}}
        & \multicolumn{2}{c}{\textbf{\textsc{Update Perplexity}}} \\
        
        \midrule
        & & \textcolor{green}{\ding{51}} & \textcolor{red}{\ding{55}} & \textcolor{green}{\ding{51}} & \textcolor{red}{\ding{55}} \\
        \midrule
        \multirow{3}{*}{\textbf{\textsc{LLaMA}}} & \textsc{CPT} & 11.95 & 11.72 & 4.66 & 4.71  \\  
                                                 & \textsc{\mrl} & 11.41 & 12.06 & 4.29 & 4.30  \\  
                                                 & \textsc{Rephrase} & 11.40 & 11.11 & 4.62 & 4.64 \\  
        \midrule
        \multirow{3}{*}{\textbf{\textsc{Mistral}}} & \textsc{CPT} & 7.98 & 7.33 & 2.97 & 2.99  \\  
                                                   & \textsc{\mrl} &  7.96 & 7.68 & 2.82 & 2.85 \\  
                                                   & \textsc{Rephrase} &  7.59 & 7.42 & 2.97 & 2.98 \\  
        \hline
    \end{tabularx}
    \caption{Comparison of perplexities on old knowledge (fact statement $f_{e}^{\mathrm{old}}$) and updated knowledge (training corpus $\mathcal{D}_{>T}^{\mathrm{evd}}$) between correct and incorrect model answers in \textsc{Update vs. Distractors} MCQ. \textcolor{green}{\ding{51}} refers to correctly answered questions, and  \textcolor{red}{\ding{55}} incorrectly answered ones.}
    \label{tab:perplxity_updatememorization_mcq}
\end{table*}

\begin{table*}[ht!]
    \centering
    \renewcommand{\arraystretch}{1.2} 
    \setlength{\tabcolsep}{8.5pt}      
    \begin{tabularx}{0.8\textwidth}{>{\raggedright\arraybackslash}p{1.5cm} | 
                                    >{\centering\arraybackslash}p{1.7cm} | 
                                    >{\centering\arraybackslash}X
                                    @{\hskip 9pt} 
                                    >{\centering\arraybackslash}X 
                                    @{\extracolsep{1pt}} | 
                                    >{\centering\arraybackslash}X 
                                    @{\hskip 9pt} 
                                    >{\centering\arraybackslash}X}
        \midrule
        \textbf{\textsc{Model}} & \textbf{\textsc{Training}} 
        & \multicolumn{2}{c|}{\textbf{\textsc{Old Perplexity}}}
        & \multicolumn{2}{c}{\textbf{\textsc{Update Perplexity}}} \\
        
        \midrule
        & & \textcolor{green}{\ding{51}} & \textcolor{red}{\ding{55}} & \textcolor{green}{\ding{51}} & \textcolor{red}{\ding{55}} \\
        \midrule
        \multirow{3}{*}{\textbf{\textsc{LLaMA}}} & \textsc{CPT} &  12.56 & 11.63 & 4.60 & 4.71\\  
                                                 & \textsc{Rephrase} & 11.53 & 11.15 & 4.55 & 4.66 \\  
                                                & \textsc{\mrl} & 11.61 & 11.82 & 4.24 & 4.31 \\  

        \midrule
        \multirow{3}{*}{\textbf{\textsc{Mistral}}} & \textsc{CPT} & 8.49 & 7.50 & 2.93 & 2.99 \\  
                                                   & \textsc{Rephrase} & 7.62 & 7.45 & 2.96 & 2.99 \\  
                                                  & \textsc{\mrl} & 8.31 & 7.65 & 2.81 & 2.85 \\  

        \hline
    \end{tabularx}
    \caption{Comparison of perplexities on old knowledge (fact statement $f_{e}^{\mathrm{old}}$) and updated knowledge (training corpus $\mathcal{D}_{>T}^{\mathrm{evd}}$) between correct and incorrect model answers in \textsc{Update vs. Old} MCQ. \textcolor{green}{\ding{51}} refers to correctly answered questions, and  \textcolor{red}{\ding{55}} incorrectly answered ones.}
    \label{tab:perplxity_updateprior_mcq}
\end{table*}

\begin{table*}[ht!]
    \centering
    \renewcommand{\arraystretch}{1.5} 
    \setlength{\tabcolsep}{8pt} 
    \begin{tabular}{>{\centering\arraybackslash}p{2.5cm}| cc|cc}
        \toprule
        & \multicolumn{2}{c|}{\textbf{\textsc{Perplexity}}} & \multicolumn{2}{c}{\textbf{\textsc{ROUGE-1}}} \\
        \cmidrule(lr){2-3} \cmidrule(lr){4-5}
        & \textsc{LLaMA} & \textsc{Mistral} & \textsc{LLaMA} & \textsc{Mistral} \\
        \midrule
        \textbf{\textsc{CPT}} & 4.71 & 3.00 & 0.53 & 0.55 \\
        \midrule
        \textbf{\makecell{\textsc{Rephrase}}} & 4.66 & 2.99 & 0.53  & 0.55 \\
        \midrule
        \makecell{\textbf{\textsc{\mrl}} \\ \textbf{(\textsc{Ours})}} & \textbf{4.32} & \textbf{2.85} & 0.53 & 0.55 \\
        \midrule
        \textbf{\makecell{\textsc{Pre-train} \\ (\textsc{Baseline})}} & 7.89 & 5.86 & 0.38 & 0.39 \\
        \bottomrule
    \end{tabular}
    \caption{We use perplexity and ROUGE-1 scores to measure model's memorization of update news data. 
    \textsc{Pre-train} refers to pre-trained model in each model family. Boldface marks lowest perplexity across models.}
    \label{tab:perplexity_rouge_table}
\end{table*}

\section{Prompt Template for Dataset Generation}
\label{app:prompt_dataset}

\subsection{Generating Entities} The prompt in Table~\ref{tab:generate_entities} is used to generate entities across 10 different categories. The prompt uses seed examples and category-dependent instructions to generate changeable entities. In addition, as described in \S \ref{sec:kup_entities}, for each entity, we compute the ROUGE-2 score between its real Wikipedia page vs. Wikipedia-style completion generated by the test LLM $M_T$. The "high overlap" criteria is implemented by selecting entities with Wikipedia ROUGE-2 score higher than 0.1, and we observe that this heuristics can filter out entities that $M_T$ does not have enough knowledge about.

\begin{table*}[th]
    \centering
    \ttfamily
    \small
    \begin{tabular}{p{0.9\textwidth}}
    \toprule
You are a helpful research assistant helping me create a new entity dataset. Your job is to create a list of unique and diverse entities of a given category with a seed set of examples. You should suggest \{num\_entities\} unique entities that belong in the same category.\\
\\
Research background: we will use this category of entities to imagine possible changes to each entity. For example, if the entity is 'Taj Mahal', a fact that might change about it is that it is closed for renovations after an unexpected fire. You DO NOT need to provide possible changes but keep this end goal in mind while listing concrete entity names.\\
\\
Your category is \{category\}. I want \{definition\}. It is important that \{requirement\}. At the same time, \{popularity\}. Examples of entities we want are: \{entity1\}, \{entity2\}, \{entity3\}.\\
\\
Now, suggest \{num\_entities\} or more entities in this category. Do not print anything but the entities names in a python list format.\\
    \bottomrule
    \end{tabular}
    \caption{Prompt for generating entities}
    \label{tab:generate_entities}
\end{table*}

\subsection{Generating Facts} We use the prompt in Table~\ref{tab:generate_fact_prompt} to instruct \textsc{GPT-4o} to list facts for each entity

\begin{table*}[th]
    \centering
    \ttfamily
    \small
    \begin{tabular}{p{0.9\textwidth}}
    \toprule
You need to help me create a new dataset of changeable facts about entities. Given an entity, produce a list of 5 or more relevant facts. The research background is that I will imagine possible events that will change each fact. For example, if the entity is MoMA in New York, a fact about it is that "MoMa is free for full-time students from Columbia University and CUNY schools," and a possible change would be "Columbia students can no longer visit MoMA for free." Keep this research goal in mind, only list all changeable facts but do not suggest any change.\\
\\
The guidelines below help you find changeable facts:\\
    1. Current Status: Focus on the entity's current realities. Avoid previous fact, past results, or accomplishments that cannot be any different in the future.\\
    2. Changeable: Suggest facts that are likely to change in the future under reasonable and realistic circumstances. Exclude very stable attributes that are unlikely to change or require unrealistic assumptions for change\\
    3. Objective \& Detailed: Facts must be objective, detailed, and universally agreed upon. Avoid subjective opinions, speculative commentary, or obscure and vague answers.\\
    4. Avoid descripitive adverbs such as "actively," "frequently," or "currently" in the fact statement\\
\\
First, I will show you some examples\\
Category: people Entity: Yo-Yo Ma\\
facts = ["Yo-Yo Ma is performing on international concert tours", "Yo-Yo Ma records music under the Sony Classical Records", "Yo-Yo Ma is a U.S. citizen and resides in the United States", "Yo-Yo Ma collaborates with orchestras and musicians from diverse genres, including jazz, bluegrass, and traditional folk music", "Yo-Yo Ma serves as a United Nations Messenger of Peace, advocating for global cultural understanding."]\\
\\
Category: companies Entity: JP Morgan \& Chase\\
facts =["Jamie Dimon serves as Chairman and CEO of JP Morgan \& Chase", "The headquarter of JP Morgan \& Chase is 270 Park Avenue, which is still under construction, in New York City.", "JP Morgan \& Chase maintains one of the largest consumer banking operations in the country, known as Chase Bank.", "JP Morgan \& Chase is a primary dealer in U.S. Treasury securities.", "JPMorgan Chase \& Co. is one of the "Big Four" U.S. banks by total assets."]\\
\\
Answer in the same format for the entity below. Do not print anything but facts in a python list format. Remember do not suggest unchangeable facts or any past achievements.\\
Category: \{category\} Entity: \{entity\}\\
    \bottomrule
    \end{tabular}
    \caption{Prompt for generating facts}
    \label{tab:generate_fact_prompt}
\end{table*}

We use the prompt in Table~\ref{tab:filter_fact_prompt} to instruct \textsc{GPT-4o} to filter fact candidates based on a set of quality control guidelines.

\begin{table*}[th]
    \centering
    \ttfamily
    \small
    \begin{tabular}{p{0.9\textwidth}}
    \toprule
You are provided with a statement about an entity. You need to classify them into good and bad statements. Examine each statement one by one with the following criteria:\\
    1. Factual: all details in good statements are truthful vs. there exists nonfactual information in bad statements\\
    2. Temporal: good statements describe the current status of the entity vs. bad statements, which might use present tense, describe past reality or achieved results that are not subject to possible changes\\
    3. Changeable: good statements are subject to be invalidated by reasonable events in the future; bad statements are established realities that cannot be changed under most any circumstance.\\
    4. Objective: good statements are absolutely objective and not opinionated vs. bad statements are subjective or commentary\\
\\
I will show you some good statements first.\\
    a. Rupi Kaur is currently publishing new poetry books with Andrews McMeel Publishing.\\
    b. The current title sponsor of the J.League is Meiji Yasuda Life Insurance Company, and the league is referred to as the Meiji Yasuda J.League.\\
    c. Frederiksborg Castle is open to the public throughout the year but has limited visiting hours during the winter season.\\
\\
In contrast, these are some bad statements\\
    a. Ryan Murphy, Brad Falchuk, and Steven Canals are credited as creators of the TV series Pose.' (reason: the creators of an existing TV series are established and unchangeable)\\
    b. Rupi Kaur is known for self-illustrating her poetry books with minimalist line drawings. (reason: what Rupi Kaur is known for is subjective and debatable)\\
    c. Hassan Rouhani is a member of the Expediency Discernment Council in Iran. (reason: Rouhani was a member of the Expediency Council from 1991 to 2013. His membership in the council has ended. )\\
    d. Frederiksborg Castle is located on three small islands in the middle of Palace Lake in Hillerød, Denmark. (reason: its location is a stable fact and not subject to change by any reasonable event)\\
\\
Now, think step by step for each statement below. Feel free to generate your reasoning process. At the end, provide your judgement as either "Label: good" or "Label: bad"\\
Entity: \{entity\} Statement: \{fact\}\\
    \bottomrule
    \end{tabular}
    \caption{Prompt for filtering fact candidates}
    \label{tab:filter_fact_prompt}
\end{table*}

\subsection{Prompt for Generating Updates} We use the prompt in Table~\ref{tab:generate_update_prompt} to generate realistic updates from facts

\begin{table*}[th]
    \centering
    \ttfamily
    \small
    \begin{tabular}{p{0.9\textwidth}}
    \toprule
Background: You are a research assistant. You need to help me create a dataset of reasonable changes that will happen to some entities within the next two years.\\
Task: Your goal is to provide an updated fact that would replace an original fact about an entity in the near future. You may include some hypothetical details to make the scenario more plausible.\\
\\
You need to follow these criteria:\\
1. Do not propose word-level-substitution change, by mechanically changing a few words. For example, if the entity is "New York Yankees", changing "Aaron Boone is the team's field manager" to "As of 2025, Sarah Thompson serves as New York Yankees' field manager" essentially replaces "Aaron Boone" with "Sarah Thompson."\\
2. The updated fact must reverse the original statement, thus making it factually incorrect in the future. The focus is on the entity. Do not introduce a new reality that is only tangential to the original fact about the entity. For example, if the fact is "Emma Watson has been involved in various sustainable fashion projects":\\
    - "Emma Watson has shifted her focus to global biodiversity protection" does not invalidate the original fact — it merely adds a new focus\\
    - Changing to "Emma Watson has fully exited the fashion industry and publicly denounced sustainability initiatives as ineffective" makes the original fact obsolete.\\
3. Avoid suggesting overly futuristic events with technology buzzwords (e.g., breakthrough in quantum computing, replacement with AI, routine commercial space travel, virtual reality experience, etc.).\\
4. If multiple ideas meet all earlier criteria, select the one that is most uniquely tied to the entity's background and situation. Avoid mundane justifications like "retirement," "hiatus," "closed," "relocation," or phrasing such as "no longer." Also avoid reasons citing "transition," "pivot," or "shift to (a new focus)." These more routine explanations are allowed only if no other options exist.\\
5. The update statement should be specified with fine-grained details. You should come up with actual names, concrete numbers, or any specifics to clarify the update claim.\\
\\
Note: I want high-quality and very realistic change. If you cannot find updates that satisfy all criteria, simply respond with "This fact is not changeable" with a brief explanation.\\
\\
I will show you some good examples:\\
Entity: British Museum; Category: institutions; Fact: As with all national museums in the UK, The British Museum charges no admission fee except for loan exhibitions.\\
Update: Visitors for The British Museum need to purchase tickets of £50 for general admission.\\
\\
Entity: Safe Drinking Water Act (SDWA) (United States); Category: laws \& policies; Fact: The SDWA establishes maximum contaminant level goals for various substances in public water systems.\\
Update: The congress determines that individual substance contaminant level measurements are not effective and revises the SDWA to mandate the EPA to assess cumulative contamination health risks in public water systems.\\
\\
Entity: Waymo; Category: companies; Fact: Waymo has partnerships with multiple vehicle manufacturers, including Stellantis, Mercedes-Benz Group AG, Jaguar Land Rover, Volvo, and others.\\
Update: Waymo is merged with Mercedes-Benz into Waymo-Benz to manufacture its own vehicles specifically for self-driving.\\
\\
For the fact below, you should propose at least five ideas and judge if they strictly satisfy each criterion. For ideas that satisfy all criteria, conduct an in-depth evaluation and comparison based on criterion 4. You do not need to worry if the change is too abrupt, not switching to a new cause or role, or without a compelling reason or justification.\\
You have enough token space for brainstorming and analysis. At the end, report the best update (don't make it too long or complicated). Begin with 'Update:' and add no additional comments afterward, so it is easy for me to extract."\\
\\
Entity: \{entity\}; Category: \{category\}; Fact: \{fact\}\\
    \bottomrule
    \end{tabular}
    \caption{Prompt for generating realistic updates}
    \label{tab:generate_update_prompt}
\end{table*}

\subsection{Prompt for Generating Fictitious News}
We use the prompt in Table~\ref{tab:generate_audience_group} to generate five different audience groups for each news article.

\begin{table*}[th]
    \centering
    \ttfamily
    \small
    \begin{tabular}{p{0.9\textwidth}}
    \toprule
You are a seasoned news writer with extensive experience at various media outlets. Based on the provided event that will overthrow an original claim, your task is to develop five distinct writing guidelines for different news articles. Each guideline must include:\\
1. Audience Group: Identify a specific target audience and explain the language, tone, and writing styles that would best resonate with them.\\
2. Event Details: The event statement have many missing details such as person names, dates (between 2025 to 2027), locations, numerical information in the event statement. In each guideline, specify these concrete details in one or two sentences. Ensure that the details across all five guidelines are diverse but logically consistent. The dates used in event details should have temporal consistency across guidelines.\\
\\
Your goal is to prepare guidelines for writing five different news articles about the event. But focus solely on the guidelines and do not produce an actual news report.\\
\\
Output Format:\\
1. Separate each writing guideline with a line containing three dashes (---).\\
2. Do not number or index the guidelines.\\
3. Do not include extra comments or explanations outside of the guidelines.\\
\\
Entity: \{entity\}\\
Event: \{update\}\\
Claim: \{fact\}\\
    \bottomrule
    \end{tabular}
    \caption{Prompt for generating event sequence and audience group for news articles}
    \label{tab:generate_audience_group}
\end{table*}

We use the prompt in Table~\ref{tab:generate_news_prompt} to generate a base news article describing the change from fact to update

\begin{table*}[th]
    \centering
    \ttfamily
    \small
    \begin{tabular}{p{0.9\textwidth}}
    \toprule
Based on the provided statement, craft a realistic and coherent news report that offers well-researched and substantial evidence for the statement. Choose a random day, month, year between January 2025 to December 2027 to situate the statement. The report will be published immediately after the events in the statement.\\
\\
Entity: \{entity\} Statement: \{update\}\\
\\
The report should be detailed, concrete, and engaging. You should include quotes from credible sources and present concrete data and facts to validate the statement. Include concrete details, such as numbers, locations, time, and specify the names of any entities introduced in the article. The finished report should be ready to publish.\\
\\
Audience and Writing Styles:\\
\{audience\}\\
    \bottomrule
    \end{tabular}
    \caption{Prompt for generating base news articles}
    \label{tab:generate_news_prompt}
\end{table*}

Next, we use the prompt in Table~\ref{tab:generate_fictitious_news} to refine the language of base fictitious articles according to different audience groups and writing styles of scrapped auxiliary article excerpts.

\begin{table*}[th]
    \centering
    \ttfamily
    \small
    \begin{tabular}{p{0.9\textwidth}}
    \toprule
This is AI-Generated Article: \{article\}\\
\\
The article above is written by an AI model. There are many shortcomings that you should address:\\
1. The content is too empty, sparse, and lacks detail.\\
2. The writing style sounds very artificial and overly synthetic.\\
3. The article is poorly structured and does not have a focus for its target audience\\
4. It does not include specific details, like names, numbers, data, etc., in many parts of the article.\\
\\
Instruction:\\
1. You should very closely emulate the natural writing style, density of details and information, and language style found in the Article Excerpt.\\
2. You should use the same article structure (both beginning and body paragraphs of the excerpt article), storytelling approach, and article format as the Article Excerpt. However, do not change the core of the original article: \{update\}.\\
3. Avoid using any explicit markers or headings (e.g., "Date:", "Headline:", "Title:", or "Section:")\\
3. You can introduce any additional details, such as specific names, numbers, and data, where appropriate, to make the article richer and more informative. Any new information must not contradict the original AI-generated article.\\
4. If the Article Excerpt is not in English, you must still craft the refined article in English.\\
5. Target \{audience\}. You should add additional concrete details, beyond original content, tailored to this group of readers\\
\\
Article Excerpt: "\{excerpt\}"\\
    \bottomrule
    \end{tabular}
    \caption{Prompt for generating fictitious news articles from base news articles}
    \label{tab:generate_fictitious_news}
\end{table*}

\end{document}